\documentclass[letterpaper]{article} 
\usepackage{aaai23}  
\usepackage{times}  
\usepackage{helvet}  
\usepackage{courier}  
\usepackage[hyphens]{url}  
\usepackage{graphicx} 
\usepackage{natbib}  
\usepackage{caption} 
\usepackage{algorithm}
\usepackage{algorithmic}

\usepackage{newfloat}
\usepackage{listings}
\DeclareCaptionStyle{ruled}{labelfont=normalfont,labelsep=colon,strut=off} 
\floatstyle{ruled}
\newfloat{listing}{tb}{lst}{}
\floatname{listing}{Listing}
\usepackage{soul}
\usepackage{amssymb}
\usepackage{pifont}
\usepackage{amsmath}
\usepackage{amsthm}
\usepackage{booktabs}
\usepackage{mathtools}
\usepackage{multirow}
\usepackage{excludeonly}
\usepackage{physics}
\usepackage{subcaption}
\usepackage{dsfont}
\newcommand{\eat}[1]{}
\newcommand{\cmark}{\ding{51}}
\newcommand{\xmark}{\ding{55}}

\newtheorem{prop}{Proposition}[section]
\newtheorem{defi}{Definition}[section]
\newtheorem{lemma}{Lemma}[section]

\title{Wiener Graph Deconvolutional Network Improves Graph Self-Supervised Learning}
\author{
    Jiashun Cheng\textsuperscript{\rm 2}, 
    Man Li\textsuperscript{\rm 2},
    Jia Li\textsuperscript{\rm 1,2}\thanks{Corresponding author.},
    Fugee Tsung\textsuperscript{\rm 1,2}
}
\affiliations{
    \textsuperscript{\rm 1}The Hong Kong University of Science and Technology (Guangzhou)\\
    \textsuperscript{\rm 2}The Hong Kong University of Science and Technology\\
    \{jchengak, mlicn\}@connect.ust.hk, \{jialee, season\}@ust.hk

}

\begin{document}

\maketitle

\begin{abstract}
Graph self-supervised learning (SSL) has been vastly employed to learn representations from unlabeled graphs. Existing methods can be roughly divided into predictive learning and contrastive learning, where the latter one attracts more research attention with better empirical performance. 
We argue that, however, predictive
models weaponed with powerful decoder could achieve
comparable or even better representation power than contrastive models. 
In this work, we propose a Wiener Graph Deconvolutional Network (WGDN), an augmentation-adaptive decoder empowered by graph wiener filter to perform information reconstruction.
Theoretical analysis proves the superior reconstruction ability of graph wiener filter.
Extensive experimental results on various datasets demonstrate the effectiveness of our approach.
\end{abstract}

\section{Introduction}\label{sec:intro}

Self-Supervised Learning (SSL), which extracts informative knowledge through well-designed pretext tasks from unlabeled data, has been extended to graph data recently due to its great success in computer vision (CV)~\cite{he2020moco} and natural language processing (NLP)~\cite{devlin2018bert}.
With regard to the objectives of pretext tasks, graph SSL can be divided into two major categories: predictive SSL and contrastive SSL~\cite{liu2021graph}.
Predictive models learn informative properties generated from graph freely via prediction tasks, while contrastive models are trained on the mutual information between  different views augmented from the original graph.
As the dominant technique, contrastive SSL has achieved state-of-the-art performance empirically~\cite{ xu2021infogcl, thakoor2022bgrl, lee2022afgrl} for graph representation learning. 
In contrast, the development of predictive SSL has lagged behind over the past few years. 

\begin{figure}[t]
    \centering
    \includegraphics[width=\linewidth]{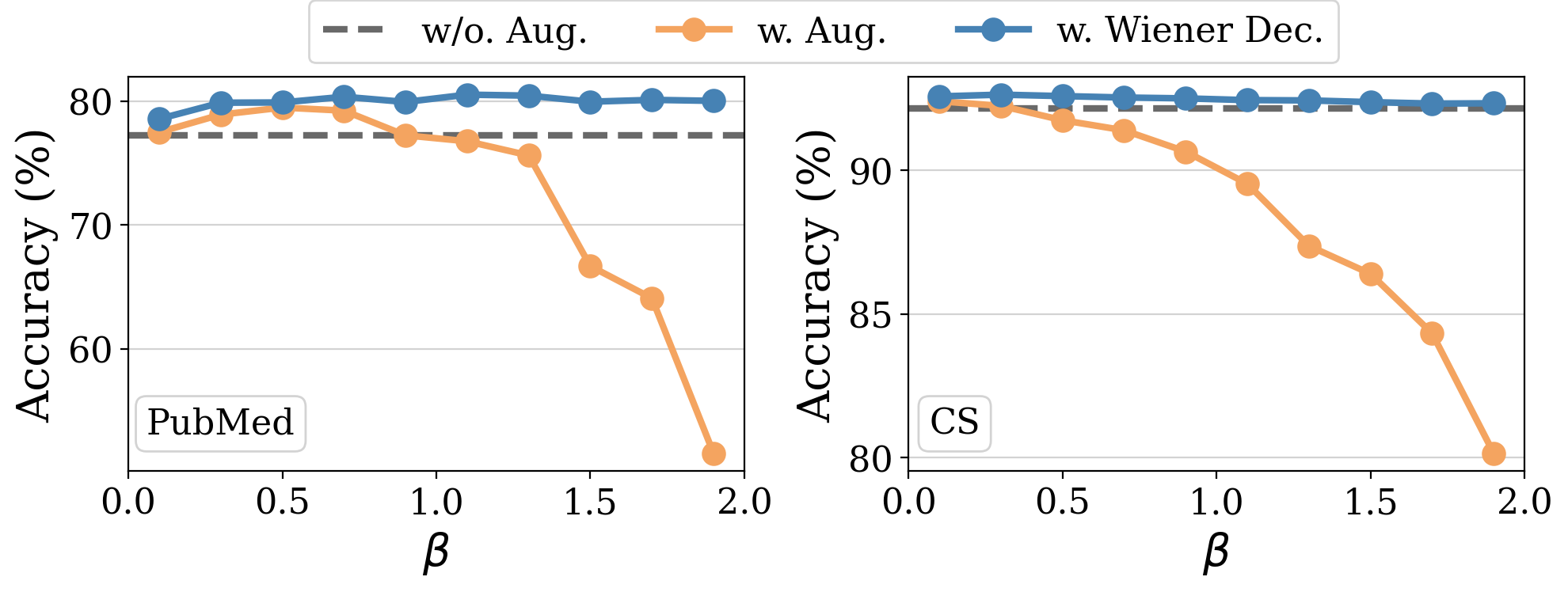}
    \caption{Comparison of different variants of GALA against latent Gaussian augmentation with magnitude $\beta$. }
    \label{fig:stable_train}
\end{figure}

Graph reconstruction is a natural self-supervision, and thus most methods in predictive SSL employ graph autoencoder (GAE) as their backbones~\cite{wang2017mgae, hu2019strategies,li2020graph}. 
The work of GraphMAE~\cite{hou2022graphmae} re-validates the potentials of reconstruction paradigm.
Despite recent advancements, \textbf{the importance of graph decoder has been largely ignored}. 
Most existing works leverage trivial decoders, such as multi-layer perceptron (MLP)~\cite{kipf2016vgae, pan2018arvga, you2020does}, which under-exploit graph topology information, and thus may lead to the degradation in learning capability.
Vanilla graph neural networks (GNNs), such as GCN~\cite{kipf2017gcn}, are inappropriate for decoding due to their Laplacian-smooth essence.
To overcome such inherent limitation of GCN, GALA~\cite{park2019gala} adopts spectral counterpart of GCN to facilitate the learning, but may take the risk of unstable learning due to its poor resilience to data augmentation (See Figure~\ref{fig:stable_train}).
GAT~\cite{velickovic2018gat} is employed as decoder in recent works including GATE~\cite{salehi2020gate} and GraphMAE~\cite{hou2022graphmae}. 
Although attention mechanism enhances model flexibility, recent work~\cite{balcilar2021analyzing} shows GAT acts like a low-pass filter and cannot well reconstruct the graph spectrum. 
As an inverse to GCN~\cite{kipf2017gcn}, \textbf{graph deconvolutional network (GDN) could be expected to further boost the performance of reconstruction}~\cite{li2021deconvolutional}, which may substantially benefit the context of representation learning. We present a summary of different decoders of predictive graph SSL in Table~\ref{tab:ssl_comp}.
Given the aforementioned observations, a natural question comes up, that is, \textit{can we improve predictive SSL by a framework with powerful decoder?}

\begin{table*}[t]
    \centering
    \begin{tabular}{c|ccccccc}
        \toprule
        \multirow{2}{*}{Model} & \multirow{2}{*}{Decoder} & Feature & Structure & Deconv. & Augmentation & Spectral & \multirow{2}{*}{Space} \\
         & & Loss & Loss & Decoder & Adaption & Kernel & \\
        \midrule
        VGAE~\cite{kipf2016vgae} & DP & - & CE & \xmark & \xmark & \xmark & $\mathcal{O}(N^2)$ \\
        ARVGA~\cite{pan2018arvga} & DP & - & CE & \xmark & \xmark & \xmark & $\mathcal{O}(N^2)$ \\
        MGAE~\cite{wang2017mgae} & MLP & MSE & - & \xmark & \xmark & \xmark & $\mathcal{O}(N)$ \\
        AttrMask~\cite{hu2019strategies} & MLP & CE & - & \xmark & \xmark & \xmark & $\mathcal{O}(N)$  \\
        GALA~\cite{park2019gala} & GNN & MSE & - & \cmark & \xmark & \xmark & $\mathcal{O}(N)$ \\
        GraphMAE~\cite{hou2022graphmae} & GNN & SCE & - & \xmark & \cmark & \xmark & $\mathcal{O}(N)$  \\
        \midrule
        WGDN & GNN & MSE & - & \cmark & \cmark & \cmark & $\mathcal{O}(N)$ \\
        \bottomrule
    \end{tabular}
    \caption{Technical components comparison within predictive SSL approaches. \textit{DP}: Non-parametric Dot Product. \textit{CE}: Cross-Entropy Error. \textit{MSE}: Mean Square Error. \textit{SCE}: Scaled-Cosine Error.}
    \label{tab:ssl_comp}
\end{table*}

Typically, a powerful decoder should at least remain effective against augmentations. Motivated by recent advancement of wiener in deep image reconstruction~\cite{dong2020deep}, we introduce the classical deconvolutional technique, wiener filter, into GDN, which is the theoretical optimum for restoring augmented signals with respect to mean square error (MSE).
We propose a GAE framework~\cite{li2020graph}, named Wiener Graph Deconvolutional Network (WGDN), which utilizes graph wiener filter to facilitate representation learning with graph spectral kernels. 
We first derive the graph wiener filter and prove its superiority in theory. 
We observe that, however, directly using the explicit graph wiener filter induces low scalability due to indispensable eigen-decomposition and may not be applicable to large-scale datasets.
Therefore, we adopt average graph spectral energy and Remez polynomial~\cite{pachon2009remez} for fast approximation. 

We evaluate the learned representation quality on two downstream tasks: node classification and graph classification. 
Empirically, our proposed WGDN achieves better results over a wide range of state-of-the-art benchmarks of graph SSL with efficient computational cost. Particularly, WGDN yields up to 1.4\% higher accuracy than runner-up model, and requires around 30\% less memory overhead against the most efficient contrastive counterpart.

\section{Related Work}\label{sec:related}

\paragraph{Graph self-supervised learning.}
According to recent surveys \cite{liu2021graph, xie2022self}, works in graph SSL can be classified into two categories: contrastive learning and predictive learning. 
Contrastive SSL attracts more attention currently due to the state-of-the-art performance on representation learning. 
Early efforts focus on the design of negative sampling and augmentation schemes, such as corruptions in DGI~\cite{velivckovic2019deep}, graph diffusion in MVGRL~\cite{hassani2020contrastive} and masking in GRACE~\cite{zhu2020grace} and GCA~\cite{zhu2021gca}.
Recent works have attempted for negative-sample-free contrastive SSL. For example, BGRL~\cite{thakoor2022bgrl} adapts BYOL~\cite{grill2020bootstrap} for graph representation learning,  CCA-SSG~\cite{zhang2021ccassg} conducts feature decorrelation, and AFGRL~\cite{lee2022afgrl} obtains positive pairs via latent space clustering. 
Despite their advancement, intricate architecture designs are required. 

As for predictive learning, predicting node features and neighborhood context is a traditional pretext task with graph autoencoder (GAE). 
For instance, VGAE~\cite{kipf2016vgae} and ARVGA~\cite{pan2018arvga} learn missing edges prediction by structural reconstruction. 
Moreover, one representative manner~\cite{you2020does} follows the perturb-then-learn strategy to predict the corrupted information, such as attribute masking ~\cite{hu2019strategies} and feature corruption ~\cite{wang2017mgae}. 
Recently, GraphMAE~\cite{hou2022graphmae} implements a masking strategy and scaled cosine error for feature reconstruction and achieves great success to match state-of-the-art contrastive SSL approaches. However, it ignores the potential benefit leveraging graph spectral theory.
In this work, we propose an augmentation-adaptive GAE framework that unleashes the power of graph spectral propagation.

\paragraph{Graph deconvolutional network.}
Regarding graph deconvolution, early research \cite{yang2018enhancing} formulates the deconvolution as a pre-processing step. 
GALA~\cite{park2019gala} performs Laplacian sharpening to recover information. 
Recent work~\cite{zhang2020graph} employs GCN~\cite{kipf2017gcn} to reconstruct node features from the latent representations. 
All these works, however, neglect the influence of augmentation.
Another GDN framework~\cite{li2021deconvolutional} is designed via a combination of inverse filters in spectral domain and denoising layers in wavelet domain, which is sub-optimal regarding signal reconstruction. 
Wiener filtering, as an alternative, executes an optimal trade-off between signal recovering and denoising. 
It has been introduced to deconvolutional networks~\cite{dong2020deep, son2017fast} for image deblurring. 
However, its effectiveness on graph structure has not been well investigated yet.

\begin{figure*}[t]
  \centering
  \includegraphics[width=0.7\linewidth]{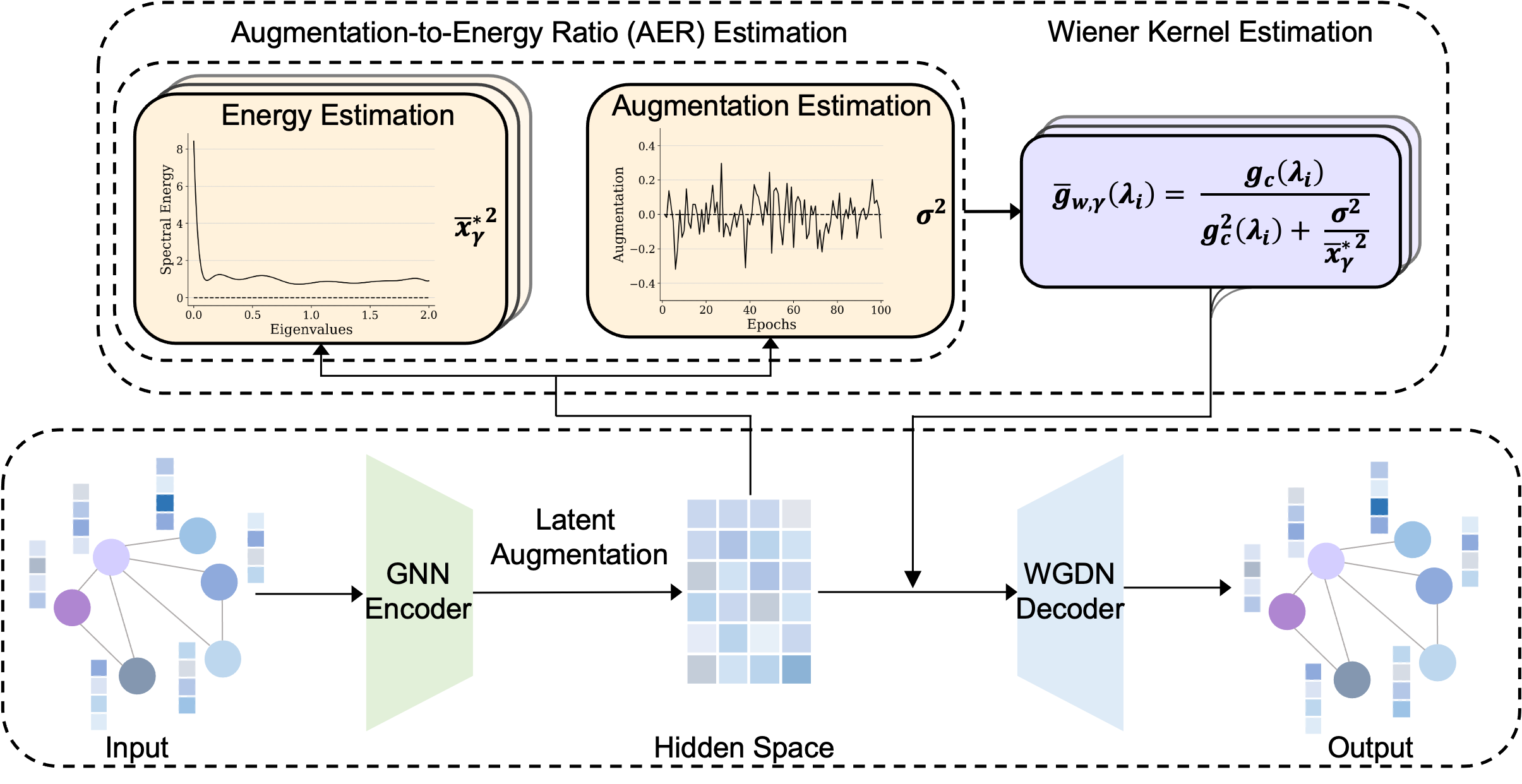}
  \caption{The autoencoder framework of WGDN for graph SSL. Given the augmented latent representations, graph wiener filter is approximated via estimating spectral energy and augmentations adaptively. With such, WGDN permits the stable feature reconstruction from the augmented latent space for representation learning.}
  \label{fig:framework}
\end{figure*}

\section{Preliminaries}\label{sec:prob}

Under a generic self-supervised graph representation learning setup, we are given an attributed graph $\mathcal{G} = (\mathcal{V}, \mathbf{A}, \mathbf{X})$ consisting of: (1) $\mathcal{V} = \{ v_{1}, v_{2}, ..., v_{N} \}$ is the set of nodes; (2) $\mathbf{A} \in \mathbb{R}^{N \times N}$  is the adjacency matrix where $\mathbf{A}_{ij} \in \{0, 1\}$ represents whether an undirected edge exists between $v_i$ and $v_j$; and (3) $\mathbf{X} \in \mathbb{R}^{N \times D}$ denotes the feature matrix. 
Our objective is to learn an autoencoder with encoder $\mathcal{E}: (\mathbb{R}^{N \times N} , \mathbb{R}^{N \times D}) \mapsto \mathbb{R}^{N \times D'}$ and decoder $\mathcal{D}: \mathbb{R}^{N \times D'} \mapsto \mathbb{R}^{N \times D}$ to produce node embedding, or graph embedding upon a pooling function. $\mathbf{H} = \mathcal{E}(\mathbf{A}, \mathbf{X}) \in \mathbb{R}^{N \times D'}$ represents the learned  embedding in low dimensional space, which can be used for various downstream tasks.

\paragraph{Graph convolution.} Convolutional operation in graph can be interpreted as a special form of Laplacian smoothing on nodes. 
From the spectral perspective, graph convolution on a signal $\mathbf{x} \in \mathbb{R}^{N}$ with a filter $g_c$ is defined as
\begin{equation} \label{eqn:conv}
\begin{aligned}
    \mathbf{h} = g_c \ast \mathbf{x} & = \mathbf{U} \text{diag}(g_c(\lambda_1), ..., g_c(\lambda_N)) \mathbf{U}^{T} \mathbf{x} \\
    & = \mathbf{U} g_c(\mathbf{\Lambda}) \mathbf{U}^{T} \mathbf{x}  = g_c(\mathbf{L}) \mathbf{x},
\end{aligned}
\end{equation}
where $\{ \lambda_i \}_{i=1}^{N}$ and $\mathbf{U}$ represent the eigenvalues and eigenvectors of normalized Laplacian matrix $\mathbf{L} = \mathbf{I} - \mathbf{D}^{-\frac{1}{2}} \mathbf{A} \mathbf{D}^{-\frac{1}{2}} = \mathbf{U} \mathbf{\Lambda} \mathbf{U}^{T}$ respectively. $\mathbf{D}$ denotes the Degree matrix. $\ast$ denotes convolutional operator.
We consider (1) GCN~\cite{kipf2017gcn}, it is a low-pass filter in spectral domain with $g_c(\lambda_i) = 1 - \lambda_i$ shown by~\cite{simgcn}; (2) GDC~\cite{klicpera2019gdc} and Heatts~\cite{li2020dirichlet}, both use heat kernel $g_c(\lambda_i) = e^{- t\lambda_i}$; (3) APPNP~\cite{gasteiger2018appnp}, it leverages personalized pagerank (PPR) kernel $g_c(\lambda_i) = \frac{\alpha}{1 - (1 - \alpha)(1 - \lambda_i)}$.

\paragraph{Graph deconvolution.} As an inverse to convolution, graph deconvolution aims to recover the input attributes given the smoothed node representation. 
From the spectral perspective, graph deconvolution  on a smoothed representation $\mathbf{h} \in \mathbb{R}^{N}$ with filter $g_d$ is defined as 
\begin{equation} \label{eqn:deconv}
\begin{aligned}
    \hat{\mathbf{x}} = g_d \ast \mathbf{h} & = \mathbf{U} \text{diag}(g_d(\lambda_1), ..., g_d(\lambda_N)) \mathbf{U}^{T} h \\
    & = \mathbf{U} g_d(\mathbf{\Lambda}) \mathbf{U}^{T} \mathbf{h}  = g_d(\mathbf{L}) \mathbf{h}.
\end{aligned}
\end{equation}

A trivial selection of $g_d$ is the inverse function of $g_c$, e.g., $g_d(\lambda_i) =\frac{1}{1 - \lambda_i}$ for GCN \cite{li2021deconvolutional}, $g_d(\lambda_i) = e^{t\lambda_i}$ for heat kernel, or $g_d(\lambda_i) = \frac{1 - (1 - \alpha)(1 - \lambda_i)}{\alpha}$ for PPR kernel.

\section{The proposed framework}\label{sec:method}

In this section, we first extend classical wiener filter to graph domain and demonstrate its superiority in reconstructing graph features. Then, we propose Wiener Graph Deconvolutional Network (WGDN), an efficient and augmentation-adaptive framework empowered by graph wiener filter.

\subsection{Wiener filter on graph} \label{sec42}

In this work, we follow the settings in previous papers~\cite{jin2019latent, cheung2021modals} and introduce additive latent augmentations in model training due to its flexible statistical characteristics, such as unbiasedness and  covariance-preserving~\cite{zhang2022costa}. Combining with the graph convolution in Eq.~\ref{eqn:conv}, augmented representation $\hat{\mathbf{h}}$ in graph is similarly defined as
\begin{equation} \label{eqn:noisegraph}
    \hat{\mathbf{h}} = \mathbf{U} g_c(\mathbf{\Lambda}) \mathbf{U}^{T} \mathbf{x} + \mathbf{\epsilon},
\end{equation}
where $\mathbf{x} \in \mathbb{R}^{N}$ denotes input features and $\mathbf{\epsilon} \in \mathbb{R}^{N}$ is assumed to be any \textit{i.i.d.} random augmentation with $\mathbb{E}[\mathbf{\epsilon}_{i}] = 0$ and  $\mathbb{VAR}[\mathbf{\epsilon}_{i}] = \sigma^2$. 
In contrast to the isolated data augmentations in graph topology and features, $\epsilon$ indirectly represents joint augmentations to both~\cite{jin2019latent}. 
Naturally, feature recovered by graph deconvolution is formulated by
\begin{equation} \label{eqn:recongraph}
    \hat{\mathbf{x}} = \mathbf{U} g_d(\mathbf{\Lambda}) g_c(\mathbf{\Lambda}) \mathbf{U}^{T} \mathbf{x} + \mathbf{U} g_d(\mathbf{\Lambda}) \mathbf{U}^T \mathbf{\epsilon}.
\end{equation}
\begin{prop} \label{prop:1}
    Let $\hat{\mathbf{x}}_{inv}$ be recovered features by inverse filter $g_d(\lambda_i) = g_c^{-1}(\lambda_i)$.
    For common low-pass filters satisfying $g_c: [0, 2] \mapsto [-1, 1]$, such as GCN, Heat and PPR, the reconstruction MSE is dominated by amplified augmentation $\text{MSE}(\hat{\mathbf{x}}_{inv}) = \mathbb{E} \norm{\mathbf{x} - \hat{\mathbf{x}}_{inv}}_{2}^{2} = \sum_{i=1}^{N} \frac{\sigma^2}{g_c^2(\lambda_i)}$.
\end{prop}

The proof is trivial and illustrated in Appendix~\ref{prop:p1} for details. Based on Proposition~\ref{prop:1}, feature reconstruction becomes unstable and even ineffective if augmentation exists.
To well utilize the power of augmentation, our goal is to stabilize the reconstruction paradigm, which resembles the classical restoration problems. 
In signal deconvolution, classical wiener filter~\cite{wienerbook} is able to produce a statistically optimal estimation of the real signals from the augmented ones with respect to MSE. 
With this regard, we are encouraged to extend wiener filter to graph domain~\cite{stationarygraph}. 
Assuming the augmentation to be independent from input features, graph wiener filter can be similarly defined by projecting MSE into graph spectral domain
\begin{equation} \label{eqn:bvmse}
\begin{aligned}
    & \text{MSE}(\hat{\mathbf{x}}) \\
    & = \mathbb{E} \norm{\hat{\mathbf{x}} - \mathbf{x}}_{2}^{2} = \mathbb{E} \norm{\mathbf{U}^T \hat{\mathbf{x}} - \mathbf{U}^T \mathbf{x}}_{2}^{2} \\
    & = \sum_{i=1}^{N} (g_d(\lambda_i) g_c(\lambda_i) - 1)^2 \mathbb{E} [x^{\ast 2}_{i}] + g_d^2(\lambda_i) \mathbb{E}[\mathbf{\epsilon}^{\ast 2}_{i}] \\
    & = \sum_{i=1}^{N} S(\lambda_i, x^{\ast}_{i}, \sigma, g_c, g_d),
\end{aligned}
\end{equation}
where $\mathbf{x}^{\ast} = \mathbf{U}^{T} \mathbf{x} = \{ x^{\ast}_1, x^{\ast}_2, ..., x^{\ast}_N \}$ and $\mathbf{\epsilon}^{\ast} = \mathbf{U}^{T} \mathbf{\epsilon} = \{ \epsilon^{\ast}_1, \epsilon^{\ast}_2, ..., \epsilon^{\ast}_N \}$ represent graph spectral projection of the input and augmentation respectively. 
We denote $\mathbb{E} [x^{\ast 2}_{i}]$ and $S(\lambda_i, x^{\ast}_{i}, \sigma, g_c, g_d)$ as the spectral energy and spectral reconstruction error of spectrum $\lambda_i$. 
Considering the convexity of Eq.~\ref{eqn:bvmse}, MSE is minimized by setting the derivative with respect to $g_d(\lambda_i)$ to zero and thus we obtain the graph wiener filter $g_w(\lambda_i)$ as
\begin{equation} \label{eqn:graphwiener}
    g_w(\lambda_i) = \frac{g_c(\lambda_i)}{g_c^2(\lambda_i) + \sigma^2/ \mathbb{E} [x^{\ast 2}_{i}]},
\end{equation}
where $\sigma^2 = \mathbb{VAR}[\mathbf{\epsilon}^{\ast}_{i}] = \mathbb{E}[\mathbf{\epsilon}^{\ast 2}_{i}]$ and $\sigma^2/\mathbb{E} [x^{\ast 2}_{i}]$ is denoted as the Augmentation-to-Energy Ratio (AER) of particular spectrum $\lambda_i$, which represents the relative magnitude of augmentation.

\begin{prop}\label{prop:2}
    Let $\hat{\mathbf{x}}_w$ be recovered features by $g_w(\lambda_i)$, where $g_w(\lambda_i)$ is a graph wiener filter, then the reconstruction MSE and variance of $\hat{\mathbf{x}}_w$ are less than $\hat{\mathbf{x}}_{inv}$. 
\end{prop}

Please refer to Appendix~\ref{prop:p2} for details. Proposition~\ref{prop:2} shows graph wiener filter has better reconstruction property than inverse filter, which promotes the resilience to latent augmentations and permits stable model training. 
We observe that, in Eq.~\ref{eqn:deconv} and~\ref{eqn:graphwiener}, eigen-decomposition is indispensable in computations of spectral energy and deconvolutional filter. 
However, in terms of scalability, an important issue for large-scale graphs is to avoid eigen-decomposition. 
Note that $\sum_{i=1}^{N} \mathbb{E} [x^{\ast 2}_{i}] = \sum_{i=1}^{N} \mathbb{E} [x^{2}_{i}]$ due to orthogonal transformation, we propose the modified graph wiener filter $\bar{g}_{w, \gamma}$ with average spectral energy $\bar{x}^{\ast 2}_{\gamma} = \gamma \cdot \bar{x}^{\ast 2} = \gamma \cdot \frac{1}{N} \sum_{i=1}^N \mathbb{E} [x^{\ast 2}_{i}]$ as
\begin{equation} \label{graphwienerm}
    \bar{g}_{w, \gamma}(\lambda_i) = \frac{g_c(\lambda_i)}{g_c^2(\lambda_i) + \sigma^2/\bar{x}^{\ast 2}_{\gamma}},
\end{equation}
where $\gamma$ is a hyper-parameter to adjust AER. As a natural extension of Proposition~\ref{prop:2}, $\bar{g}_{w, \gamma}$ owns the following proposition.

\begin{prop} \label{prop:3}
    Let $\hat{\mathbf{x}}_{w, \gamma}$ be the recovered features by modified graph wiener filter $\bar{g}_{w, \gamma} (\lambda_i)$, then the variance of $\hat{\mathbf{x}}_{w, \gamma}$ is less than $\hat{\mathbf{x}}_{inv}$. In spectral domain, given two different $\gamma_1$, $\gamma_2$ such that $\mathbb{E} [x^{\ast 2}_{i}] \leq \bar{x}^{\ast 2}_{\gamma_1} \leq \bar{x}^{\ast 2}_{\gamma_2}$, the spectral reconstruction error $S(\lambda_i, x^{\ast}_i, \sigma, g_c, \bar{g}_{w, \gamma_1}) \leq S(\lambda_i, x^{\ast}_i, \sigma, g_c, \bar{g}_{w, \gamma_2}) \leq S(\lambda_i, x^{\ast}_i, \sigma, g_c, g_c^{-1})$.
\end{prop}

Please refer to Appendix~\ref{prop:p3} for details. Proposition~\ref{prop:3} demonstrates that $\bar{g}_{w, \gamma}$ attends to spectral reconstructions over different ranges of spectra, depending on the selection of $\gamma$. 
The graph wiener kernel $\mathbf{D}_{\gamma} = \mathbf{U} \bar{g}_{w, \gamma}(\mathbf{\Lambda}) \mathbf{U}^{T}$ can also be reformatted as matrix multiplication
\begin{equation} \label{eqn:wienermatrix}
    \mathbf{D}_{\gamma} = \mathbf{U} (g_c^2(\mathbf{\Lambda}) + \frac{\sigma^2}{\bar{x}^{\ast 2}_{\gamma}} \mathbf{I})^{-1} g_c(\mathbf{\Lambda}) \mathbf{U}^T.
\end{equation}

Note that $g_c$ can be arbitrary function and support of $\lambda_i$ is restricted to [0, 2], we adopt Remez polynomial~\cite{pachon2009remez} to approximate $\bar{g}_{w, \gamma}(\lambda_i)$, which mitigates the need of eigen-decomposition and matrix inversion in Eq.~\ref{eqn:wienermatrix}. 

\begin{defi}[\textit{\textbf{Remez Polynomial Approximation}}] \label{defi:1}
    Given an arbitrary continuous function $\zeta(t)$ on $t \in [a, b]$, the Remez polynomial approximation  for $\zeta(t)$ is defined as 
    \begin{equation}
    \begin{aligned}
        p_{K}(t) \coloneqq \sum_{k=0}^{K} c_{k} t^{k},
    \end{aligned}
    \end{equation}
    where coefficients $c_{0}, \dots, c_{K}$ and leveled error $e$ are obtained by resolving linear system
    \begin{equation} \label{eqn:remezpoints}
    \begin{aligned}
        \zeta(t_{j}) = p_{K}(t_{j}) + (-1)^{j} e,
    \end{aligned}
    \end{equation}
    where $\{ t_{j} \}_{j=0}^{K+1}$ are interpolation points within [a, b].
\end{defi}

\begin{lemma}
     If interpolation points $\{ t_{j} \}_{j=0}^{K+1}$ are Chebyshev nodes, the interpolation error $|\zeta(t) - p_{K}(t)|$ of Remez polynomial $p_{K}(t)$ is minimized. 
\end{lemma}

The proof is trivial and illustrated in detail as Corollary 8.11 in~\cite{burden2011numerical}. Following Definition~\ref{defi:1}, the $K^{\text{th}}$ order Remez approximation of $\mathbf{D}_{\gamma}$ is formulated as
\begin{equation} \label{eqn:wienerremez}
\begin{aligned}
    \mathbf{D}_{\gamma} & = \mathbf{U} p_{K}(\mathbf{\Lambda}) \mathbf{U}^{T} = \sum_{k=0}^{K} c_{k, \gamma} \mathbf{L}^{k},
\end{aligned}
\end{equation}
where $\mathbf{D}_{\gamma}$ is approximated adaptively in each epoch.

\subsection{Wiener graph deconvolutional network} \label{sec44}

\paragraph{Graph encoder.} To incorporate both graph features $\mathbf{X}$ and structure $\mathbf{A}$ in a unified framework, we employ $M$ layers of graph convolution neural network as our graph encoder. For $m = 0, ..., M - 1$,
\begin{equation} \label{eqn:encoder}
    \mathbf{H}^{(m+1)} = \phi(g_c(\mathbf{L}) \mathbf{H}^{(m)} \mathbf{W}^{(m)}),
\end{equation}
where $\mathbf{H}^{(0)} = \mathbf{X}$, $\phi$ is the activation function such as PReLU and $g_c(\lambda_i) = 1 - \lambda_i$ as GCN~\cite{kipf2017gcn}, $g_c(\lambda_i) = e^{- t\lambda_i}$ as heat kernel or $g_c(\lambda_i) = \frac{\alpha}{1 - (1 - \alpha)(1 - \lambda_i)}$ as PPR kernel.

\paragraph{Representation augmentation.} For simplicity, Gaussian noise is employed as latent augmentations to the node embedding generated by the last layer encoder 
\begin{equation} \label{eqn:latent_aug}
    \mathbf{\hat{H}}^{(M)} = \mathbf{H}^{(M)} + \beta \mathbf{E},
\end{equation}
where $\mathbf{E} = \{ \mathbf{\epsilon}_{1}, ..., \mathbf{\epsilon}_{N}\}$, $\mathbf{\epsilon_{i}} \sim N(\mathbf{0}, \sigma^2_{P} \mathbf{I})$, $\sigma^2_{P} =  \mathbb{VAR}[\mathbf{H}^{(M)}]$ and $\beta$ is a hyper-parameter to adjust the magnitude of augmentations.

\paragraph{Graph wiener decoder.} The decoder aims to recover original features given the augmented representation $\mathbf{\hat{H}}$. Our previous analysis demonstrates the superiority of wiener kernel to permit reconstruction-based representation learning with augmented latent space. Considering the properties of spectral reconstruction error from Proposition~\ref{prop:3}, we symmetrically adopt $M$ layers of graph deconvolution as the decoder, where each layer consists of $q$ channels of graph wiener kernels. For $m = 1, ..., M$ and $i = 1, ..., q$,
\begin{equation} \label{eqn:wienerdecoder}
\begin{aligned}
    \mathbf{Z}^{(m-1)}_{i} & = \phi(\mathbf{D}_{\gamma_i}^{(m)} \mathbf{\hat{H}}^{(m)} \mathbf{W}^{(m)}_{i}), \\
    \mathbf{\hat{H}}^{(m-1)} & = \text{AGG}([\mathbf{Z}^{(m-1)}_{1}, ..., \mathbf{Z}^{(m-1)}_{q}]),
\end{aligned}
\end{equation}
where $\mathbf{\hat{X}} = \mathbf{\hat{H}}^{(0)}$ and AGG($\cdot$) is aggregation function such as summation. 
Note that the actual value of $\bar{x}^{\ast 2}$ and $\sigma^2$ of $\mathbf{D}_{\gamma_i}^{(m)}$ are unknown, we estimate $\bar{x}^{\ast 2}$ following its definition and leverage neighboring information for $\sigma^2$ estimation. Further details are presented in Appendix~\ref{model:detail_arch}.

\paragraph{Optimization and inference.} Our model is optimized following the convention of reconstruction-based SSL, which is simply summarized as
\begin{equation}
    \mathcal{L} = ||\mathbf{X} - \hat{\mathbf{X}}||_{F}.
\end{equation}
For downstream applications, we treat the fully trained $\mathbf{H}^{(M)}$ as the final node embedding. For graph-level tasks, we adopt a non-parametric graph pooling (readout) function $\mathcal{R}$, e.g. MaxPooling, to generate graph representation $\mathbf{h}_{g} = \mathcal{R}(\mathbf{H}^{(M)})$.

\paragraph{Complexity analysis.} The most intensive computational cost of our proposed method is kernel approximation in Eq.~\ref{eqn:wienerremez}. Note that kernel approximation is a simple $K^{\text{th}}$ order polynomial of graph convolution. By sparse-dense matrix multiplication, graph convolution can be efficiently implemented, which take $O(K|E|)$~\cite{kipf2017gcn} for a graph with $|E|$ edges.

\section{Experiments}\label{sec:exper}

In this section, we investigate the benefit of our proposed approach by addressing the following questions: 

\textbf{Q1.} Does WGDN outperform self-supervised and semi-supervised counterparts?

\textbf{Q2.} Do the key components of WGDN contribute to representation learning?

\textbf{Q3.} Can WGDN be more efficient than competitive baselines?

\textbf{Q4.} How do the hyper-parameters impact the performance of our proposed model?

\begin{table*}[ht]
    \centering
    \begin{tabular}{c|c|ccccc}
        \toprule
        & Model & PubMed & Computers & Photo & CS & Physics \\
        \midrule
        \multirow{13}{*}{Self-supervised} & Node2Vec & 66.6 $\pm$ 0.9 & 84.39 $\pm$ 0.08 &  89.67 $\pm$ 0.12 & 85.08 $\pm$ 0.03 & 91.19 $\pm$ 0.04 \\
        & DeepWalk + Feat. & 74.3 $\pm$ 0.9 &  86.28 $\pm$ 0.07 & 90.05 $\pm$ 0.08 & 87.70 $\pm$ 0.04 & 94.90 $\pm$ 0.09 \\
        \cmidrule{2-7}
        & GAE & 72.1 $\pm$ 0.5 & 85.27 $\pm$ 0.19 & 91.62 $\pm$ 0.13 & 90.01 $\pm$ 0.71 & 94.92 $\pm$ 0.07 \\
        & GALA & 75.9 $\pm$ 0.4 & 87.61 $\pm$ 0.06 & 91.27 $\pm$ 0.12 & 92.48 $\pm$ 0.07 & 95.23 $\pm$ 0.04 \\
        & GDN & 76.4 $\pm$ 0.2 & 87.67 $\pm$ 0.17 & 92.84 $\pm$ 0.07 & 92.93 $\pm$ 0.18 & 95.22 $\pm$ 0.05 \\
        \cmidrule{2-7}
        & DGI & 76.8 $\pm$ 0.6 & 83.95 $\pm$ 0.47 & 91.61 $\pm$ 0.22 & 92.15 $\pm$ 0.63 & 94.51 $\pm$ 0.52 \\
        & MVGRL & 80.1 $\pm$ 0.7 & 87.52 $\pm$ 0.11 & 91.74 $\pm$ 0.07 & 92.11 $\pm$ 0.12 & 95.33 $\pm$ 0.03 \\
        & GRACE & 80.5 $\pm$ 0.4 & 86.25 $\pm$ 0.25 & 92.15 $\pm$ 0.24 & 92.93 $\pm$ 0.01 & 95.26 $\pm$ 0.02 \\
        & GCA & 80.2 $\pm$ 0.4 & 88.94 $\pm$ 0.15 & 92.53 $\pm$ 0.16 & 93.10 $\pm$ 0.01 & 95.73 $\pm$ 0.03 \\
        & BGRL$^{\ast}$ & 79.8 $\pm$ 0.4 & \underline{89.70 $\pm$ 0.15} & 93.37 $\pm$ 0.21 & 93.51 $\pm$ 0.10 & 95.28 $\pm$ 0.06 \\
        & AFGRL$^{\ast}$ & 79.9 $\pm$ 0.3 & 89.58 $\pm$ 0.45 & \underline{93.61 $\pm$ 0.20} & \underline{93.56 $\pm$ 0.15} & \underline{95.74 $\pm$ 0.10} \\
        & CCA-SSG$^{\ast}$ & \underline{81.0 $\pm$ 0.3} & 88.15 $\pm$ 0.35 & 93.25 $\pm$ 0.21 & 93.31 $\pm$ 0.16 & 95.59 $\pm$ 0.07 \\
        \cmidrule{2-7}
        & WGDN & \textbf{81.9 $\pm$ 0.4} & \textbf{89.72 $\pm$ 0.48} & \textbf{93.89 $\pm$ 0.31} & \textbf{93.67 $\pm$ 0.14} & \textbf{95.76 $\pm$ 0.11} \\
        \midrule
        \multirow{2}{*}{Supervised} & GCN & 79.1 $\pm$ 0.3 & 86.51 $\pm$ 0.54 & 92.42 $\pm$ 0.22 & 93.03 $\pm$ 0.31 & 95.65 $\pm$ 0.16 \\
        & GAT & 79.0 $\pm$ 0.3 & 86.93 $\pm$ 0.29 & 92.56 $\pm$ 0.35 & 92.31 $\pm$ 0.24 & 95.47 $\pm$ 0.15 \\
        \bottomrule
    \end{tabular}
    \caption{Node classification accuracy of all compared methods. The best and runner up models in self-supervised learning are highlighted in boldface and underlined.}
    \label{tab:exp_node}
    
    \bigskip
    
    \small
    \begin{tabular}{c|c|cccccc}
        \toprule
        & Model & IMDB-B & IMDB-M & PROTEINS & COLLAB & DD & NCI1 \\
        \midrule
        \multirow{11}{*}{Self-supervised} & WL & 72.30 $\pm$ 3.44 & 46.95 $\pm$ 0.46 & 72.92 $\pm$ 0.56 & 79.02 $\pm$ 1.77 & 79.43 $\pm$ 0.55 & 80.01 $\pm$ 0.50 \\
        & DGK & 66.96 $\pm$ 0.56 & 44.55 $\pm$ 0.52 & 73.30 $\pm$ 0.82 & 73.09 $\pm$ 0.25 & - & 80.31 $\pm$ 0.46 \\
        \cmidrule{2-8}
        & Graph2Vec & 71.10 $\pm$ 0.54 & 50.44 $\pm$ 0.87 & 73.30 $\pm$ 2.05 & - & - & 73.22 $\pm$ 1.81 \\
        & MVGRL & 74.20 $\pm$ 0.70 & 51.20 $\pm$ 0.50 & - & -  & - & - \\
        & InfoGraph & 73.03 $\pm$ 0.87 & 49.69 $\pm$ 0.53 & 74.44 $\pm$ 0.31 & 70.65 $\pm$ 1.13 & 72.85 $\pm$ 1.78 & 76.20 $\pm$ 1.06 \\
        & GraphCL & 71.14 $\pm$ 0.44 & 48.58 $\pm$ 0.67 &  74.39 $\pm$ 0.45 & 71.36 $\pm$ 1.15 & 78.62 $\pm$ 0.40 & 77.87 $\pm$ 0.41 \\
        & JOAO & 70.21 $\pm$ 3.08 & 49.20 $\pm$ 0.77 & 74.55 $\pm$ 0.41 & 69.50 $\pm$ 0.36 & 77.32 $\pm$ 0.54 & 78.07 $\pm$ 0.47 \\
        & SimGRACE & 71.30 $\pm$ 0.77 & - & \underline{75.35 $\pm$ 0.09} & 71.72 $\pm$ 0.82 & 77.44 $\pm$ 1.11 & 79.12 $\pm$ 0.44 \\
        & InfoGCL & 75.10 $\pm$ 0.90 & 51.40 $\pm$ 0.80 & - & 80.00 $\pm$ 1.30 & - & 80.20 $\pm$ 0.60 \\
        & GraphMAE & \underline{75.52 $\pm$ 0.66} & \underline{51.63 $\pm$ 0.52} & 75.30 $\pm$ 0.39 & \underline{80.32 $\pm$ 0.46} & \underline{78.86 $\pm$ 0.35} & \underline{80.40 $\pm$ 0.30} \\
        \cmidrule{2-8}
        & WGDN & \textbf{75.76 $\pm$ 0.20} & \textbf{51.77 $\pm$ 0.55} & \textbf{76.53 $\pm$ 0.38} & \textbf{81.76 $\pm$ 0.24} & \textbf{79.54 $\pm$ 0.51} & \textbf{80.70 $\pm$ 0.39} \\
        \midrule
        \multirow{2}{*}{Supervised} & GCN & 74.0 $\pm$ 3.4 & 51.9 $\pm$ 3.8 & 76.0 $\pm$ 3.2 & 79.0 $\pm$ 1.8 & 75.9 $\pm$ 2.5 & 80.2 $\pm$ 2.0\\
        & GIN & 75.1 $\pm$ 5.1 & 52.3 $\pm$ 2.8 & 76.2 $\pm$ 2.8 & 80.2 $\pm$ 1.9 & 75.3 $\pm$ 2.9 & 82.7 $\pm$ 1.7\\
        \bottomrule
    \end{tabular}
    \caption{Graph classification accuracy of all compared methods.}
    \label{tab:exp_graph}
\end{table*}

\subsection{Experimental setup} \label{sec:exper_setup}

\paragraph{Datasets.} We conduct experiments on both node-level and graph-level representation learning tasks with benchmark datasets across different scales and domains, including PubMed~\cite{sen2008collective}, Amazon Computers, Photo~\cite{shchur2018pitfalls}, Coauthor CS, Physics~\cite{shchur2018pitfalls}, and IMDB-B, IMDB-M, PROTEINS, COLLAB, DD, NCI1 from TUDataset~\cite{morris2020tudataset}. 
Detailed statistics are presented in Table~\ref{tab:dataset_node} and Table~\ref{tab:dataset_graph} of Appendix~\ref{exper:spec}.

\paragraph{Baselines.} We compare WGDN against representative models from the following five different categories: 
(1) traditional models including Node2Vec~\cite{grover2016node2vec}, Graph2Vec~\cite{narayanan2017graph2vec},  DeepWalk~\cite{perozzi2014deepwalk}, 
(2) graph kernel models including Weisfeiler-Lehman sub-tree kernel (WL)~\cite{shervashidze2011wl}, deep graph kernel (DGK)~\cite{yanardag2015dgk}, 
(3) predictive SSL models including GAE~\cite{kipf2016vgae},  GALA~\cite{park2019gala}, GDN~\cite{li2021deconvolutional}, GraphMAE~\cite{hou2022graphmae}, 
(4) contrastive SSL models including DGI~\cite{velivckovic2019deep}, MVGRL~\cite{hassani2020contrastive}, GRACE~\cite{zhu2020grace}, GCA~\cite{zhu2021gca}, BGRL~\cite{thakoor2022bgrl}, AFGRL~\cite{lee2022afgrl}, CCA-SSG~\cite{zhang2021ccassg}, InfoGraph~\cite{sun2019infograph}, GraphCL~\cite{you2020graphcl}, JOAO~\cite{you2021joao}, SimGRACE~\cite{xia2022simgrace}, InfoGCL~\cite{xu2021infogcl} and 
(5) semi-supervised models including GCN~\cite{kipf2017gcn}, GAT~\cite{velickovic2018gat} and GIN~\cite{xu2018gin}.

\paragraph{Evaluation protocol.} We closely follow the evaluation protocol in recent SSL researches. For node classification, the node embedding is fed into a logistic regression classifier~\cite{velivckovic2019deep}. 
We run 20 trials with different seeds and report the mean classification accuracy with standard deviation. 
For graph classification, we feed the graph representation into a linear SVM, and report the mean 10-fold cross-validation accuracy with standard deviation after 5 runs~\cite{xu2021infogcl}. Please refer to Appendix~\ref{exper:spec:eval} for further details.

\paragraph{Experiment settings.} We use the official implementations for all baselines in node classification and follow the suggested hyper-parameter settings, whereas graph classification results are obtained from original papers if available. 
For spectral filter, we consider heat kernel $g_c(\lambda_i) = e^{-t\lambda_i}$ with diffusion time $t = 1$ and PPR kernel $g_c(\lambda_i) = \frac{\alpha}{1 - (1 - \alpha)(1 - \lambda_i)}$ with teleport probability $\alpha = 0.2$.
In node classification training, we use the public split for PubMed and follow 10/10/80\% random split for the rest.
Further details of model configurations (e.g., hyper-parameters selection) can be found in Appendix~\ref{exper:spec:hyper}.

\subsection{Performance comparison (Q1)} \label{sec:exper_q1}

The node classification performances are reported in Table~\ref{tab:exp_node}. We find that WGDN outperforms the predictive SSL methods by a large margin over all datasets. For fair comparisons, we report the best results of recent methods using diffusion kernels (denoted with $^\ast$). WGDN performs competitively with contrastive SSL methods, achieving state-of-the-art performances across all datasets. For instance, our model WGDN is able to improve by a margin up to 0.9\% on accuracy over the most outstanding contrastive counterpart CCA-SSG on PubMed. Moreover, when compared to semi-supervised models, WGDN consistently generates better performance than both GCN and GAT. 

Table~\ref{tab:exp_graph} lists the graph classification performance across various methods. We observe that our approach achieves state-of-the-art results compared to existing SSL baselines in all datasets. Besides, WGDN outperforms the best kernel methods up to a large margin. Even when compared to semi-supervised models, our model achieves the best results in 4 out of 6 datasets and the gaps for the rest are relatively minor.

In brief, our model consistently achieves comparable performance with the cutting-edge SSL and semi-supervised methods across node-level and graph-level tasks. 
Particularly, the significant improvements demonstrate the effectiveness of WGDN in boosting the learning capability under GAE framework.

\subsection{Effectiveness of key components (Q2)} \label{sec:exper_q2}

To validate the benefit of introducing graph wiener decoder, we conduct ablation studies on node and graph classification tasks with five datasets that exhibit distinct characteristics (e.g., citation, social and bioinformatics). 
For clarity, WGDN-A and WGDN-W are denoted as the models
removing augmentation or substituting graph wiener decoder with inverse decoder. WGDN-AW is the plain model without both components.
Specifically, heat kernel is selected as the backbone of encoder for node-level datasets, and we adopt PPR kernel for graph-level datasets. 

\begin{figure}[t]
    \centering
    \includegraphics[width=\linewidth]{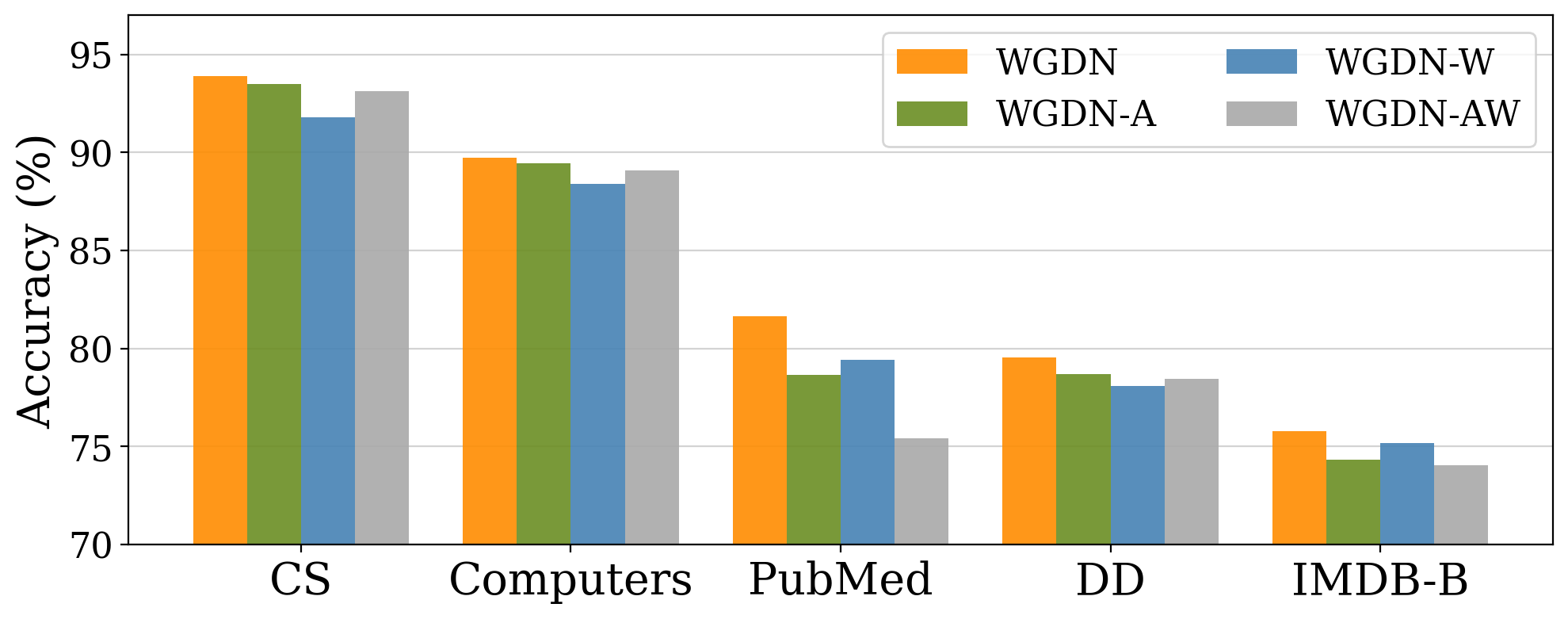}
    \caption{Ablation study of graph wiener decoder. Complete model consistently boosts model performance across different datasets.}
    \label{fig:ablation}
\end{figure}

The results are illustrated in Figure~\ref{fig:ablation}, from which we make several observations. 
\textbf{(1)} WGDN-W may underperform WGDN-AW. This observation validates that deterministic inverse decoder is ill-adapted to augmented latent space and may lead to degraded learning quality, which is consistent with our theoretical analysis in Section~\ref{sec42}. 
\textbf{(2)} Compared with WGDN-AW, WGDN-A improves model performance across all datasets, which suggests that graph wiener decoder is able to benefit representation learning even without augmentation.
\textbf{(3)} The performance of WGDN is significantly higher than other counterparts. For instance, WGDN has a relative improvement up to 6\% over WGDN-AW on PubMed. It can be concluded that the graph wiener decoder allows the model to generate more semantic embedding from the augmented latent space.

\subsection{Efficiency analysis (Q3)} \label{sec:exper_q3}

To evaluate the computational efficiency, we compare the training speed and GPU overhead of WGDN against BGRL and GraphMAE on datasets of different scales, including Computers and OGBN-Arxiv~\cite{hu2020ogb}.
For fair comparisons, we set the embedding size of all models as 512 and follow their suggested hyper-parameters settings. 
It is evident from Table~\ref{tab:exp_efficiency} that the memory requirement of WGDN is significantly reduced up to 30\% compared to BGRL, the most efficient contrastive benchmark. 
In addition, as WGDN is a GAE framework without computationally expensive add-on, its computational cost is shown to be comparable to GraphMAE.
Considering that memory is usually the bottleneck in graph-based applications, WGDN demonstrates a practical advantage when limited resources are available.

\begin{table}[h]
    \centering
    \begin{tabular}{cccc}
    \toprule
    Dataset & Model & Steps/Second & Memory\\
    \midrule
    \multirow{3}{*}{Computers} & BGRL & 17.27 & 3.01 GB \\
     & GraphMAE & 19.47 & \textbf{2.03 GB} \\
     & WGDN & 19.62 & 2.20 GB \\
    \midrule
    \multirow{3}{*}{OGBN-Arxiv} & BGRL & 2.52 & 9.74 GB \\
     & GraphMAE & 3.13 & 8.01 GB \\
     & WGDN & 3.16 & \textbf{7.35 GB} \\
    \bottomrule
    \end{tabular}
    \caption{Comparison of computational efficiency on benchmark datasets.}
    \label{tab:exp_efficiency}
\end{table}

\begin{figure}[t]
    \centering
    \begin{subfigure}{.5\linewidth}
        \centering
        \includegraphics[width=\linewidth]{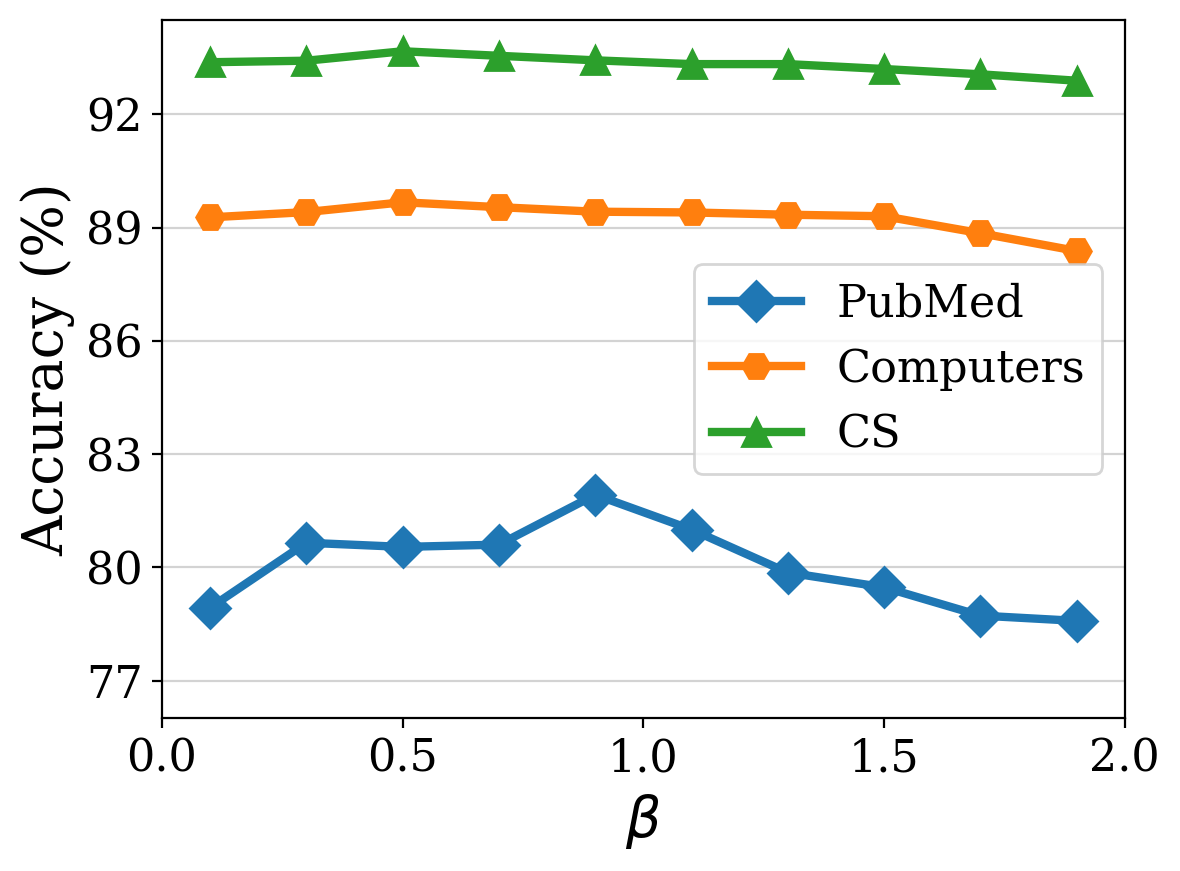}
    \end{subfigure}%
    \begin{subfigure}{.5\linewidth}
        \centering
        \includegraphics[width=\linewidth]{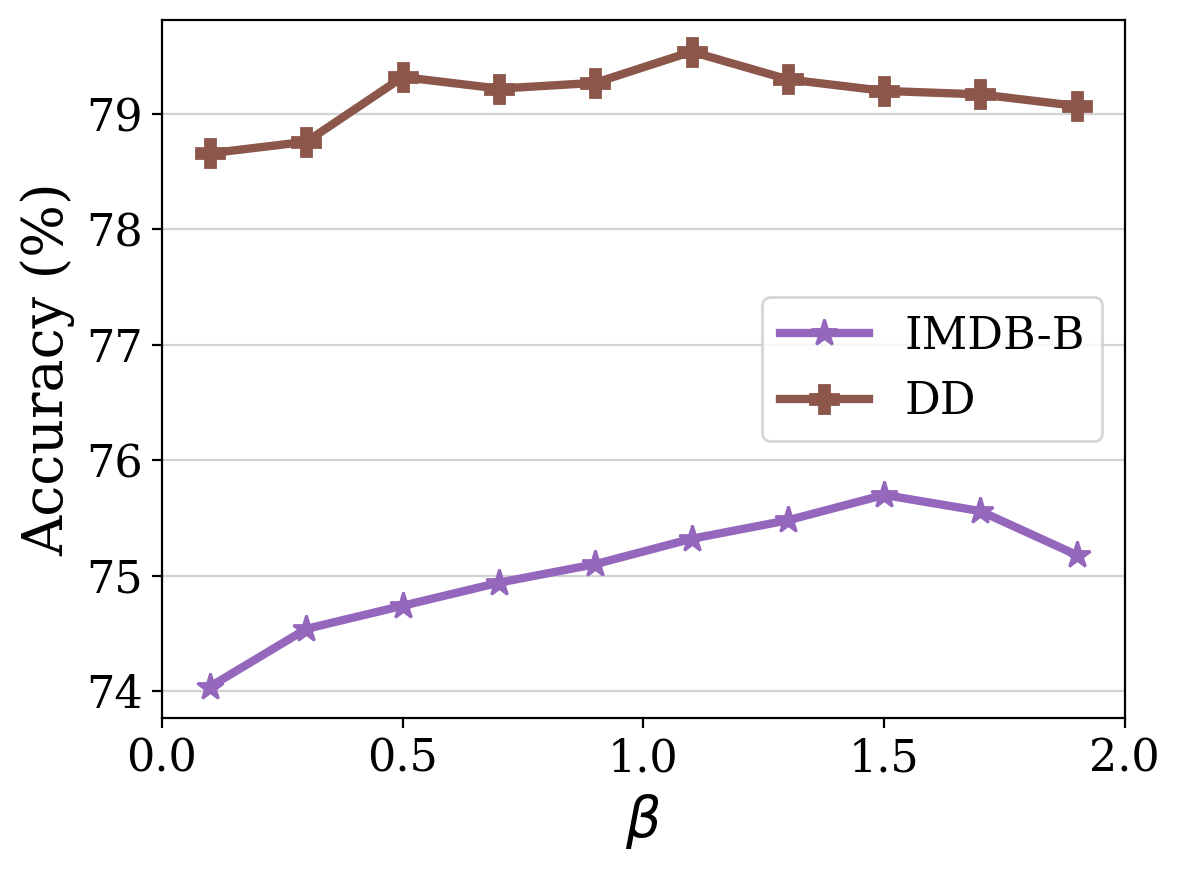}
    \end{subfigure}
    \caption{Downstream tasks performance versus varied augmentation magnitude $\beta$ in training.}
    \label{fig:sens}
\end{figure}

\begin{table}[t]
    \centering
    \small
    \begin{tabular}{cccc}
        \toprule
        Filter & GCN & Heat & PPR \\
        \midrule
        PubMed & 80.2 (0.019) & \textbf{81.9 (0.011)} & 81.4 (0.013) \\
        Computers & 89.03 (0.417) & \textbf{89.72 (0.375)} & 89.59 (0.405) \\
        CS & 92.48 (0.263) & \textbf{93.67 (0.241)} & 92.75 (0.245) \\
        \midrule
        IMDB-B & 75.46 (0.102) & 75.71 (0.098) & \textbf{75.76 (0.093)} \\
        DD & 79.29 (0.118) & 79.36 (0.104) & \textbf{79.54 (0.074)} \\
        \bottomrule
    \end{tabular}
    \caption{Performance and training loss of WGDN with different convolution filter $g_c$.}
    \label{tab:exp_filter}
\end{table}

\subsection{Hyper-parameter analysis (Q4)} \label{sec:exper_q4}

\paragraph{Magnitude of Augmentation $\beta$.}
It is expected that introducing adequate augmentation enriches the sample distribution in the latent space, which contributes to learning more expressive representations. 
Figure~\ref{fig:sens} shows that the classification accuracy generally reaches the peak and drops gradually when the augmentation size $\beta$ increases, which aligns with our intuition.
We also observe that the optimal augmentation magnitudes are relatively smaller for node-level datasets, which may be related to the semantics level of graph features. 
Input features of graph-level datasets are less informative and latent distribution may still preserve with stronger augmentations. 
Besides, the stable trend further verifies that graph wiener decoder is well adapted to augmentation in representation learning. 

\paragraph{Convolution filter $g_c$.} 
Table~\ref{tab:exp_filter} shows the influence of different convolution filters. 
It is observed that diffusion-based WGDN outperforms its trivial version with GCN filter across different applications. 
Specifically, heat kernel generates better results in node classification and PPR kernel is more suitable for graph-level tasks. We conjecture that sparse feature information may be better compressed via propagation with PPR kernel.
In addition, we also find that training loss of diffusion models is consistently lower. 
Both phenomena indicate that the superior information aggregation and powerful reconstruction of diffusion filters jointly contribute to learning a more semantic representation.

\section{Conclusion and future work}\label{sec:conc}

In this paper, we propose Wiener Graph Deconvolutional Network (WGDN), a predictive self-supervised learning framework for graph-structured data. 
We introduce graph wiener filter and theoretically validate its superior reconstruction ability to facilitate reconstruction-based representation learning.
By leveraging graph wiener decoder, our model can efficiently learn graph embedding with augmentation. 
Extensive experimental results on various datasets demonstrate that  WGDN achieves competitive performance over a wide range of self-supervised and semi-supervised counterparts. 

\section*{Acknowledgements}\label{sec:ack}

This work was supported by the Hong Kong RGC General Research Funds 16216119, Foshan HKUST Projects FSUST20-FYTRI03B, in part by NSFC Grant 62206067 and Guangzhou-HKUST(GZ) Joint Funding Scheme.

\bibliography{aaai23}


\clearpage
\appendix
\section*{Technical Appendix}
In the technical appendix, we provide further details for the proofs, model architectures and experiment.

\section{Proof of Proposition~\ref{prop:1}}\label{prop:p1}
\begin{proof} 
By substitution, $\mathbf{\hat{x}}_{inv} = \mathbf{x} + \mathbf{U} g_d(\mathbf{\Lambda}) \mathbf{U}^T \mathbf{\epsilon}$. Note that MSE is reduced to
    \begin{equation}
    \begin{aligned}
        \mathbb{E} \norm{\mathbf{x} - \mathbf{\hat{x}}_{inv}}_{2}^{2} = \mathbb{E} \norm{\mathbf{U} g_d(\mathbf{\Lambda}) \mathbf{U}^T \mathbf{\epsilon}}_{2}^{2} = \sum_{i=1}^{N} \frac{\sigma^2}{g_c^2(\lambda_i)}.
    \end{aligned}
    \end{equation}
    Regarding the condition $g_c: [0, 2] \mapsto [-1, 1]$, we take GCN where $g_c(\lambda_i) = 1 - \lambda_i$ as a representative example. By substitution, we have
    \begin{equation}
    \begin{aligned}
        \mathbb{E} \norm{\mathbf{x} - \mathbf{\hat{x}}_{inv}}_{2}^{2} = \sum_{i=1}^{N} \frac{\sigma^2}{(1 - \lambda_i)^2},
    \end{aligned}
    \end{equation}
and $\frac{1}{(1-\lambda_i)^2} \to \infty$ when $\lambda_i \to 1$.
\end{proof}

\section{Proof of Proposition~\ref{prop:2}}\label{prop:p2}
\begin{proof} 
    By the definition of $g_w(\lambda_i)$
    \begin{equation}
    \begin{aligned}
        \mathbb{E} \norm{\mathbf{x} - \mathbf{\hat{x}}_w}_{2}^{2} & = \sum_{i=1}^{N} \frac{\sigma^2 (g_c^2(\lambda_i) + \sigma^2/\mathbb{E} [x^{\ast 2}_{i}])}{(g_c^2(\lambda_i) + \sigma^2/\mathbb{E} [x^{\ast 2}_{i}])^2} \\
        & = \sum_{i=1}^{N} \frac{\sigma^2}{g_c^2(\lambda_i) + \sigma^2/\mathbb{E} [x^{\ast 2}_{i}]} \\
        & \leq \sum_{i=1}^{N} \frac{\sigma^2}{g_c^2(\lambda_i)}.
    \end{aligned}
    \end{equation}
    Plugging in the inverse filter, we can obtain
    \begin{equation}
    \begin{aligned}
        & \sum_{i=1}^{N} \mathbb{VAR}[\mathbf{\hat{x}}_{inv, i}] \\
        & = \text{Tr}(\mathbb{COV}[\mathbf{\hat{x}}_{inv}]) \\
        & = \text{Tr}(\mathbb{COV}[\mathbf{U} \mathbf{x}^{\ast}] + \mathbb{COV}[\mathbf{U} g_d(\mathbf{\Lambda}) \mathbf{\epsilon}^{\ast}]) \\
        & = \sum_{i=1}^{N} (\mathbb{VAR}[\mathbf{x}^{\ast}_{i}] + \frac{\sigma^2}{g_c^2(\lambda_i)}),
    \end{aligned}
    \end{equation}
    where $\text{Tr}$ represents the matrix trace. Similarly, variance of $\mathbf{\hat{x}}_w$ is convoluted by $g_w(\lambda_i)$
    \begin{equation}
    \begin{aligned}
        & \sum_{i=1}^{N} \mathbb{VAR}[\mathbf{\hat{x}}_{w, i}] \\
        & = \text{Tr}(\mathbb{COV}[\mathbf{\hat{x}}_{w}]) \\
        & = \sum_{i=1}^{N} [\frac{g_c^2(\lambda_i) }{g_c^2(\lambda_i) + \sigma^2/\mathbb{E} [x^{\ast 2}_{i}]}]^2 (\mathbb{VAR}[\mathbf{x}^{\ast}_{i}] + \frac{\sigma^2}{g_c^2(\lambda_i)}) \\
        & \leq \sum_{i=1}^{N} (\mathbb{VAR}[\mathbf{x}^{\ast}_{i}] + \frac{\sigma^2}{g_c^2(\lambda_i)}).
    \end{aligned}
    \end{equation}
\end{proof}

\section{Proof of Proposition~\ref{prop:3}}\label{prop:p3}
\begin{proof}
    Similar to Appendix~\ref{prop:p2}, variance of $\mathbf{\hat{x}}_{w, \gamma}$ is reduced to
    \begin{equation}
    \begin{aligned}
        & \sum_{i=1}^{N} \mathbb{VAR}(\mathbf{\hat{x}}_{w, \gamma, i}) \\
        & = \text{Tr}(\mathbb{COV}(\mathbf{\hat{x}}_{w, \gamma})) \\
        & = \sum_{i=1}^{N} [\frac{g_c^2(\lambda_i) }{g_c^2(\lambda_i) + \sigma^2/x^{\ast 2}_{\gamma}}]^2 (\mathbb{VAR}[\mathbf{x}^{\ast}_{i}] + \frac{\sigma^2}{g_c^2(\lambda_i)}) \\
        & \leq \sum_{i=1}^{N} (\mathbb{VAR}[\mathbf{x}^{\ast}_{i}] + \frac{\sigma^2}{g_c^2(\lambda_i)}).
    \end{aligned}
    \end{equation}
    For the specific spectrum $\lambda_i$ where $\mathbb{E} [x^{\ast 2}_{i}] \leq \bar{x}^{\ast 2}_{\gamma}$ holds, the spectral reconstruction error satisfies
    \begin{equation}
    \begin{aligned}
        & S(\lambda_i, x^{\ast}_{i}, \sigma, g_{c}, \bar{g}_{w, \gamma}) \\ 
        & = \frac{1}{(g_c^2(\lambda_i) + \sigma^2/\bar{x}^{\ast 2}_{\gamma})^2} [\sigma^2 (g_c^2(\lambda_i) + \frac{\sigma^2}{\bar{x}^{\ast 2}_{\gamma}} \cdot \frac{\mathbb{E} [x^{\ast 2}_{i}]}{\bar{x}^{\ast 2}_{\gamma}})] \\
        & \leq \frac{1}{(g_c^2(\lambda_i) + \sigma^2/\bar{x}^{\ast 2}_{\gamma})^2} [\sigma^2 (g_c^2(\lambda_i) + \frac{\sigma^2}{\bar{x}^{\ast 2}_{\gamma}})] \\
        & = \frac{\sigma^2}{g_c^2(\lambda_i) + \sigma^2/\bar{x}^{\ast 2}_{\gamma}} \\
        & \leq \frac{\sigma^2}{g_c^2(\lambda_i)} = S(\lambda_i, x^{\ast}_{i}, \sigma, g_{c}, g_{c}^{-1}).
    \end{aligned}
    \end{equation}
    Note that the second derivative of spectral reconstruction error $S(\lambda_i, x^{\ast}_{i}, \sigma, g_{c}, g_{d})$ with respect to $g_{d}(\lambda_i)$ is
    \begin{equation}
    \begin{aligned}
        & \frac{\partial^2}{\partial g_{d}^{2}(\lambda_i)} S(\lambda_i, x^{\ast}_{i}, \sigma, g_{c}, g_{d}) \\
        & = 2(g_{c}^{2}(\lambda_i) \mathbb{E} [x^{\ast 2}_{i}] + \sigma^2) \geq 0,
    \end{aligned}
    \end{equation}
    thus, $S(\lambda_i, x^{\ast}_{i}, \sigma, g_{c}, g_{d})$ is a convex function. By Eq.~\ref{eqn:graphwiener}, $g_{w}(\lambda_i)$ is the solution for global minimum. By convexity, for any filter $g_{d}(\lambda_i)$, the value of $S(\lambda_i, x^{\ast}_{i}, \sigma, g_{c}, g_{d})$ is greater when distance to global minimizer $|g_{d}(\lambda_i) - g_{w}(\lambda_i)|$ is larger. Considering  $\bar{g}_{w, \gamma}(\lambda_i)$, it can be reduced to
    \begin{equation}
    \begin{aligned}
        & |\bar{g}_{w, \gamma}(\lambda_i) - g_{w}(\lambda_i)| \\ 
        & = |\frac{g_{c}(\lambda_i)}{g_c^2(\lambda_i) + \sigma^2/\bar{x}^{\ast 2}_{\gamma}} - \frac{g_{c}(\lambda_i)}{g_{c}^2(\lambda_i) + \sigma^2/\mathbb{E} [x^{\ast 2}_{i}]}|.
    \end{aligned}
    \end{equation}
    Given the condition that $x^{\ast 2}_{i} \leq \bar{x}^{\ast 2}_{\gamma_1} \leq \bar{x}^{\ast 2}_{\gamma_2}$, we can conclude that
    \begin{equation}
    \begin{aligned}
        |\bar{g}_{w, \gamma_1}(\lambda_i) - g_{w}(\lambda_i)| \leq |\bar{g}_{w, \gamma_2}(\lambda_i) - g_{w}(\lambda_i)|.
    \end{aligned}
    \end{equation}
    Therefore, $S(\lambda_i, x^{\ast}_i, \sigma, g_c, \bar{g}_{w, \gamma_1}) \leq S(\lambda_i, x^{\ast}_i, \sigma, g_c, \bar{g}_{w, \gamma_2}) \leq S(\lambda_i, x^{\ast}_i, \sigma, g_c, g_c^{-1})$ holds.
\end{proof}

\begin{table*}[ht]
    \centering
    \begin{tabular}{c|ccccc}
        \toprule
        Model & PubMed & Computers & Photo & CS & Physics \\
        \midrule
        BGRL$_{\text{N}}$ & 78.6 $\pm$ 0.7 & 89.49 $\pm$ 0.21 & 93.01 $\pm$ 0.20 & 93.15 $\pm$ 0.12 & 95.14 $\pm$ 0.06 \\
        BGRL$_{\text{H}}$ & 79.8 $\pm$ 0.4 & 89.70 $\pm$ 0.15 & 93.37 $\pm$ 0.21 & 93.51 $\pm$ 0.10 & 95.24 $\pm$ 0.09 \\
        BGRL$_{\text{P}}$ & 79.4 $\pm$ 0.5 & 89.63 $\pm$ 0.17 & 93.15 $\pm$ 0.21 & 93.42 $\pm$ 0.12 & 95.28 $\pm$ 0.06 \\
        \midrule
        AFGRL$_{\text{N}}$ & 79.5 $\pm$ 0.2 & 88.91 $\pm$ 0.37 & 92.96 $\pm$ 0.25 & 93.17 $\pm$ 0.15 & 95.54 $\pm$ 0.09 \\
        AFGRL$_{\text{H}}$ & 79.9 $\pm$ 0.3 & 89.58 $\pm$ 0.45 & 93.61 $\pm$ 0.20 & 93.56 $\pm$ 0.15 & 95.74 $\pm$ 0.10 \\
        AFGRL$_{\text{P}}$ & 79.7 $\pm$ 0.3 & 89.33 $\pm$ 0.37 & 93.06 $\pm$ 0.27 & 93.53 $\pm$ 0.14 & 95.71 $\pm$ 0.10 \\
        \midrule
        CCA-SSG$_{\text{N}}$ & 80.5 $\pm$ 0.3 & 87.35 $\pm$ 0.30 & 92.38 $\pm$ 0.33 & 93.31 $\pm$ 0.16 & 95.14 $\pm$ 0.07 \\
        CCA-SSG$_{\text{H}}$ & 81.0 $\pm$ 0.3 & 88.15 $\pm$ 0.35 & 93.25 $\pm$ 0.25 & - & 95.59 $\pm$ 0.07 \\
        CCA-SSG$_{\text{P}}$ & 80.7 $\pm$ 0.3 & 87.27 $\pm$ 0.46 & 92.79 $\pm$ 0.25 & - & 95.12 $\pm$ 0.13 \\
        \midrule
        $\text{WGDN}_{\text{N}}$ & 80.2 $\pm$ 0.4 & 89.03 $\pm$ 0.46 & 92.26 $\pm$ 0.37 & 92.48 $\pm$ 0.12 & 95.33 $\pm$ 0.02 \\
        $\text{WGDN}_{\text{H}}$ & \textbf{81.9 $\pm$ 0.4} & \textbf{89.72 $\pm$ 0.48} & \textbf{93.89 $\pm$ 0.31} & \textbf{93.67 $\pm$ 0.17} & \textbf{95.76 $\pm$ 0.11} \\
        $\text{WGDN}_{\text{P}}$ & 81.4 $\pm$ 0.3 & 89.59 $\pm$ 0.45 & 92.96 $\pm$ 0.19 & 92.75 $\pm$ 0.21 & 95.40 $\pm$ 0.23 \\
        \bottomrule
    \end{tabular}
    \caption{Node classification accuracy with different encoding propagation backbones.}
    \label{tab:exp_backbones}
\end{table*}

\begin{table*}[ht]
    \centering
    \begin{tabular}{cccccc}
        \toprule
        Model & PubMed & Comp & CS & IMDB-B & DD \\
        \midrule
        AFGRL & 79.9 $\pm$ 0.3 & 89.58 $\pm$ 0.45 & 93.56 $\pm$ 0.15 & 75.07 $\pm$ 0.58 & 78.58 $\pm$ 0.44 \\
        GraphMAE & 81.1 $\pm$ 0.4 & 89.53 $\pm$ 0.31 & 93.51 $\pm$ 0.13 & 75.52 $\pm$ 0.66 & 78.86 $\pm$ 0.35 \\
        \midrule
        WGDN-DE & 80.9 $\pm$ 0.6 & 89.55 $\pm$ 0.36 & 93.56 $\pm$ 0.31 & 75.42 $\pm$ 0.15 & 79.24 $\pm$ 0.40 \\ 
        WGDN-DN & 81.3 $\pm$ 0.5 & 89.49 $\pm$ 0.34 & 93.53 $\pm$ 0.31 & 75.52 $\pm$ 0.17 & 79.31 $\pm$ 0.32 \\ 
        WGDN & \textbf{81.9 $\pm$ 0.4} & \textbf{89.72 $\pm$ 0.48} & \textbf{93.89 $\pm$ 0.31} & \textbf{75.76 $\pm$ 0.20} & \textbf{79.54 $\pm$ 0.51} \\
        \bottomrule
    \end{tabular}
    \caption{Performance of WGDN against different augmentation methods.}
    \label{tab:exp_augmentation}
\end{table*}

\section{Details of Model Architecture}
\label{model:detail_arch}

\paragraph{AER estimation.} Let $\mathbf{\hat{H}}^{(m)}$ denotes the input of $m$-th layer decoder, the average spectral energy $\bar{x}^{\ast 2}_{\gamma_i}$ in $\mathbf{D}_{\gamma_i}^{(m)}$ is estimated following  $\sum_{i=1}^{N} \mathbb{E} [x^{\ast 2}_{i}] = \sum_{i=1}^{N} \mathbb{E} [x^{2}_{i}] = \sum_{i=1}^{N} {\mathbb{E} [x_{i}]}^2 + \mathbb{VAR} [x_{i}]$. Specifically, 
\begin{equation}
    \bar{x}^{\ast 2}_{\gamma_i} = \frac{\gamma_i}{ND'} \big( \norm{\mathbf{\hat{H}}^{(m)}}_F^2 + \norm{\mathbf{\hat{H}}^{(m)} - \frac{1}{N} \mathds{1} \mathbf{\hat{H}}^{(m)}}_F^2 \big),
\end{equation}
where $\mathds{1} \in \mathbb{R}^{N \times D'}$ is the all ones matrix and $D'$ is the size of hidden space. The augmentation variance $\sigma^2$ is estimated by considering its neighborhood as
\begin{equation}
    \sigma^2 = \frac{1}{ND'} \norm{\mathbf{\hat{H}}^{(m)} - \mathbf{D}^{-1} \mathbf{A} \mathbf{\hat{H}}^{(m)}}_{F}^{2},
\end{equation}
where $\mathbf{A}$ and $\mathbf{D}$ are adjacency matrix and degree matrix.

\paragraph{Skip connection.} To learn more expressive representations, skip connection is considered in training phase for some cases as it transmits aggregated information to create 'hard' examples for decoder, which may encourage encoder to compress more useful knowledge. 
If skip connection is implemented, we let $\mathbf{\hat{H}}^{(m)}_{d} = \mathbf{\hat{H}}^{(m)}$ for clarity. For $m=1, .., M - 1$, we augment the output of the $m$-layer encoder, denoted as $\mathbf{\hat{H}}^{(m)}_{e}$, by
\begin{equation}
    \mathbf{\hat{H}}^{(m)}_{e} = \mathbf{H}^{(m)} + \beta \mathbf{E}^{(m)},
\end{equation}
where $\mathbf{E}^{(m)} = \{ \mathbf{\epsilon}_{1}^{(m)}, ..., \mathbf{\epsilon}_{N}^{(m)}\}$, $\mathbf{\epsilon}_{i}^{(m)} \sim N(\mathbf{0}, \sigma_{P}^{2 \, (m)} \mathbf{I})$, $\sigma_{P}^{2 \, (m)} = \mathbb{VAR}[\mathbf{H}^{(m)}]$ and $\beta$ is same hyper-parameter in Eq.~\ref{eqn:latent_aug}. To avoid learning trivial representation, both augmented representations are fed into the same decoder as
\begin{equation}
\begin{aligned}
    \mathbf{Z}^{(m-1)}_{i, s} & = \phi(\mathbf{D}_{\gamma_i, s}^{(m)} \mathbf{\hat{H}}^{(m)}_{s} \mathbf{W}^{(m)}_{i}), \\
    \mathbf{\hat{H}}^{(m-1)}_{s} & = \text{AGG}([\mathbf{Z}^{(m-1)}_{1, s}, ..., \mathbf{Z}^{(m-1)}_{q, s}]), 
\end{aligned}
\end{equation}
where $s = \{e, d\}$ represents the source. The final representation of $m$-layer decoder is obtained by averaging the intermediate embeddings,
\begin{equation}
    \mathbf{\hat{H}}^{(m-1)} = \text{AVG}(\mathbf{\hat{H}}^{(m-1)}_{e}, \mathbf{\hat{H}}^{(m-1)}_{d}).
\end{equation}

\paragraph{Normalization.} For graph classification, we apply batch normalization~\cite{ioffe2015batch} right before activation function for all layers except the final prediction layer.

\section{More Experimental Results}

\subsection{Experiment on Encoder Backbones}

To have a fair comparison with contrastive methods, we compare WGDN with the three state-of-the-art baselines using spectral kernel in node classification task. 
For clarity, models with GCN, heat and PPR kernel are denoted with subscript $_\text{N}$, $_\text{H}$ and $_\text{P}$ respectively. 

From Table~\ref{tab:exp_backbones}, it is observed that WGDN still achieves better performance over baselines trained with spectral kernels.
In terms of CS, CCA-SSG employs MLP as its decoder because GNN decoders worsen model performance under its experiment settings. 
In addition, utilizing spectral propagation consistently boosts the learning capability, which aligns with our motivation pertaining to the potential of graph spectral kernel.
Particularly, for node classification, heat kernel may be a better option, as it brings the greatest improvement regardless of datasets and model types. 

\begin{table*}[ht]
    \centering
    \begin{tabular}{c|cccccccc}
        \toprule
        Dataset & Cora & CiteSeer & PubMed & Computers & Photo & CS & Physics & OGBN-Arxiv \\
        \midrule
        \# Nodes & 2,708 & 3,327 & 19,717 & 13,752 & 7,650 & 18,333 & 34,493 & 169,343 \\
        \# Edges & 10,556 & 9,104 & 88,648 & 491,722 & 238,162 & 163,788 & 495,924 & 1,166,243 \\
        \# Classes & 7 & 6 & 3 & 10 & 8 & 15 & 5 & 40 \\
        \# Features & 1,433 & 3,703 & 500 & 767 & 745 & 8,415 & 8,415 & 128 \\
        \midrule
        Augmentation $\beta$ & 0.9 & 1.0 & 1.0 & 0.4 & 0.5 & 0.5 & 0.2 & 0.8 \\
        Hidden Size & 512 & 512 & 1024 & 512 & 512 & 512 & 512 & 768 \\
        Epoch & 100& 100 & 300 & 1000 & 1000 & 150 & 100 & 120 \\
        Filter $g_c(\lambda_i)$ & PPR & PPR & Heat & Heat & Heat & Heat & Heat & PPR \\
        Aggregation & Max & Sum & Max & - & - & - & - & - \\
        Last Activation & \cmark & - & - & \cmark & - & - & \cmark & - \\
        Skip Connection & \cmark & - & \cmark & - & - & \cmark & \cmark & - \\
        \bottomrule
    \end{tabular}
    \caption{Summary of datasets and hyper-parameter configuration for node classification task.}
    \label{tab:dataset_node}
\end{table*}

\begin{table*}[ht]
    \centering
    \begin{tabular}{c|cccccc}
        \toprule
        Dataset & IMDB-B & IMDB-M & PROTEINS & COLLAB & DD & NCI1\\
        \midrule
        \# Graphs & 1,000 & 1,500 & 1,113 & 5,000 & 1,178 & 4,110 \\
        \# Avg. Nodes & 19.77 & 13.00 & 39.06 & 74.49 &  284.32 & 29.87 \\
        \# Avg. Edges & 193.06 & 65.94 & 72.82 & 2457.78 &  715.66 & 32.30 \\
        \# Classes & 2 & 3 & 2 & 3 & 2 & 2 \\
        \midrule
        Augmentation $\beta$ & 1.5 & 1.5 & 1.0 & 1.0 & 1.0 & 0.5 \\
        Learning Rate & 0.0001 & 0.0001 & 0.0001 & 0.0001 & 0.0001 & 0.0005 \\
        Batch Size & 32 & 32 & 32 & 32 & 32 & 16 \\
        Epoch & 100 & 100 & 10 & 20 & 40 & 500 \\
        Filter $g_c(\lambda_i)$ & PPR & PPR & PPR & PPR &  PPR & Heat \\
        Aggregation & Max & Avg & Avg & - & Avg & - \\
        Pooling & Avg & Avg & Max & Max & Sum & Max \\
        Skip Connection & \cmark & \cmark & \cmark & \cmark & \cmark & - \\
        \bottomrule
    \end{tabular}
    \caption{Summary of datasets and hyper-parameter configuration for graph classification task.}
    \label{tab:dataset_graph}
\end{table*}

\subsection{Experiment on Different Augmentations}

Although the graph wiener decoder is derived based on latent augmentations, a powerful decoder should adapt to general augmentation techniques. We conduct experiments with five representative datasets with drop-edge and drop-node, which are denoted as WGDN-DE and WGDN-DN respectively. Table~\ref{tab:exp_augmentation} shows that different variants of our framework still achieve comparable performance with competitive baselines on both node and graph classification. However, general augmentations hardly satisfy the distribution assumption, resulting in inaccurate noise estimation and performance degradation.

\subsection{Experiments on Additional Datasets}

To evaluate the effectiveness of our proposed method, three more commonly used datasets are considered. For fair comparisons, we conducted hyper-parameter search to finetune the most competitive baselines. We report the results in Table~\ref{tab:exp_common_extra} and find that WGDN still achieves state-of-the-art performances in 2 out of 3 datasets.

\begin{table}[ht]
    \centering
    \begin{tabular}{cccc}
        \toprule
        Model & Cora & CiteSeer & OBGN-Arxiv \\
        \midrule
        BGRL & 82.8 $\pm$ 0.5 & 71.3 $\pm$ 0.8 & 71.64 $\pm$ 0.12 \\
        AFGRL & 81.7 $\pm$ 0.4 & 71.2 $\pm$ 0.4 & 71.39 $\pm$ 0.16 \\
        CCA-SSG & 83.8 $\pm$ 0.5 & \textbf{73.1 $\pm$ 0.3} & 71.24 $\pm$ 0.20 \\
        \midrule
        WGDN & \textbf{84.2 $\pm$ 0.6} & 72.2 $\pm$ 1.1 & \textbf{71.76 $\pm$ 0.23} \\
        \bottomrule
    \end{tabular}
    \caption{Node classification accuracy on additional commonly used datasets.}
    \label{tab:exp_common_extra}
\end{table}

\section{Detailed Experimental Setup} \label{exper:spec}

\subsection{Evaluation Protocol} \label{exper:spec:eval}

For node classification tasks, all the resulted embedding from well-trained unsupervised models are frozen. We use Glorot initialization~\cite{glorot2010understanding} to initialize model parameters and all downstream models are trained for 300 epochs by Adam optimizer~\cite{kingma2014adam} with a learning rate 0.01. We run 20 trials and keep the model with the highest performance in validation set of each run as final. 
For graph classification tasks, linear SVM is fine-tuned with grid search on \textit{C} parameter from $\{10^{-3}, 10^{-2}, ..., 1, 10\}$.

\subsection{Hyper-parameter Specifications} \label{exper:spec:hyper}

By default, wiener graph decoder is implemented with multiple channels with $q = 3$ and $\gamma =  [0.1, 1, 10]$. Otherwise, single channel is used with $\gamma = 1$. Aggregation function is applied only in the scenarios of multiple channels.
For wiener kernel approximation, the polynomial order $K$ of GCN kernel is 9 while others are 2.
All models are initialized using Glorot initialization~\cite{glorot2010understanding} and optimized by Adam optimizer~\cite{kingma2014adam}. 

For node classification, we set the learning rate as 0.001. The number of layers is 3 for OGBN-Arxiv and set as 2 for the remaining datasets. To enhance the representation power of embedding, no activation function is employed in the last layer of encoder in some cases. 
For graph classification, the dimension of hidden embedding is set to 512. The number of layer is set to 3 for IMDB-M as well as PROTEINS and 2 for the rest.
The detailed hyper-parameter configurations of each dataset are illustrated in Table~\ref{tab:dataset_node} and~\ref{tab:dataset_graph}.

\subsection{Baselines Implementations}

For node classification, we use the official implementation of BGRL, AFGRL and CCA-SSG and follow the suggested hyper-parameter settings for reproduction. For fair comparison in the scenarios of spectral kernel implementation, we conduct hyper-parameter search for them and select the model with best results in validation set as final.

For graph classification, we report the previous results in the public papers. For dataset DD, we generate the result of GraphMAE with its source code and hyper-parameter searching.

\subsection{Computational Hardware}

We use the machine with the following configurations for all model training and evaluation.

\begin{itemize}
    \item OS: Ubuntu 20.04.1 LTS
    \item CPU: Intel(R) Xeon(R) Silver 4114 CPU @ 2.20GHz
    \item GPU: NVIDIA Tesla V100, 32GB 
\end{itemize}

\end{document}